\documentclass{article}

\PassOptionsToPackage{numbers, compress}{natbib}
\usepackage[final]{neurips_2024}

\usepackage[utf8]{inputenc} 
\usepackage[T1]{fontenc}    
\usepackage{hyperref}       
\usepackage{url}            
\usepackage{booktabs}       
\usepackage{amsfonts}       
\usepackage{nicefrac}       
\usepackage{microtype}      
\usepackage{xcolor}         
\usepackage{graphicx}
\usepackage{float}
\usepackage{wrapfig}
\usepackage{amsmath}
\usepackage{stmaryrd}

\title{Self-Attention Limits Working Memory Capacity of Transformer-Based Models}

\author{%
  Dongyu Gong\\
  Yale University\\
  New Haven, CT 06510 \\
  \texttt{dongyu.gong@yale.edu} \\
  \And
  Hantao Zhang \\
  Yale University\\
  New Haven, CT 06510 \\
  \texttt{hantao.zhang@yale.edu} \\
}

\begin{document}

\maketitle

\begin{abstract}
  Recent work on Transformer-based large language models (LLMs) has revealed striking limits in their working memory capacity, similar to what has been found in human behavioral studies. Specifically, these models' performance drops significantly on \textit{N}-back tasks as \textit{N} increases. However, there is still a lack of mechanistic interpretability as to why this phenomenon would arise. Inspired by the executive attention theory from behavioral sciences, we hypothesize that the self-attention mechanism within Transformer-based models might be responsible for their working memory capacity limits. To test this hypothesis, we train vanilla decoder-only transformers to perform \textit{N}-back tasks and find that attention scores gradually aggregate to the \textit{N}-back positions over training, suggesting that the model masters the task by learning a strategy to pay attention to the relationship between the current position and the \textit{N}-back position. Critically, we find that the total entropy of the attention score matrix increases as \textit{N} increases, suggesting that the dispersion of attention scores might be the cause of the capacity limit observed in \textit{N}-back tasks. Our findings thus offer insights into the shared role of attention in both human and artificial intelligence. Moreover, the limitations of the self-attention mechanism revealed in the current study could inform future efforts to design more powerful model architectures with enhanced working memory capacity and cognitive capabilities.
\end{abstract}

\section{Introduction}

\begin{figure*}[t]
    \begin{center}
        \includegraphics[width=0.95\linewidth]{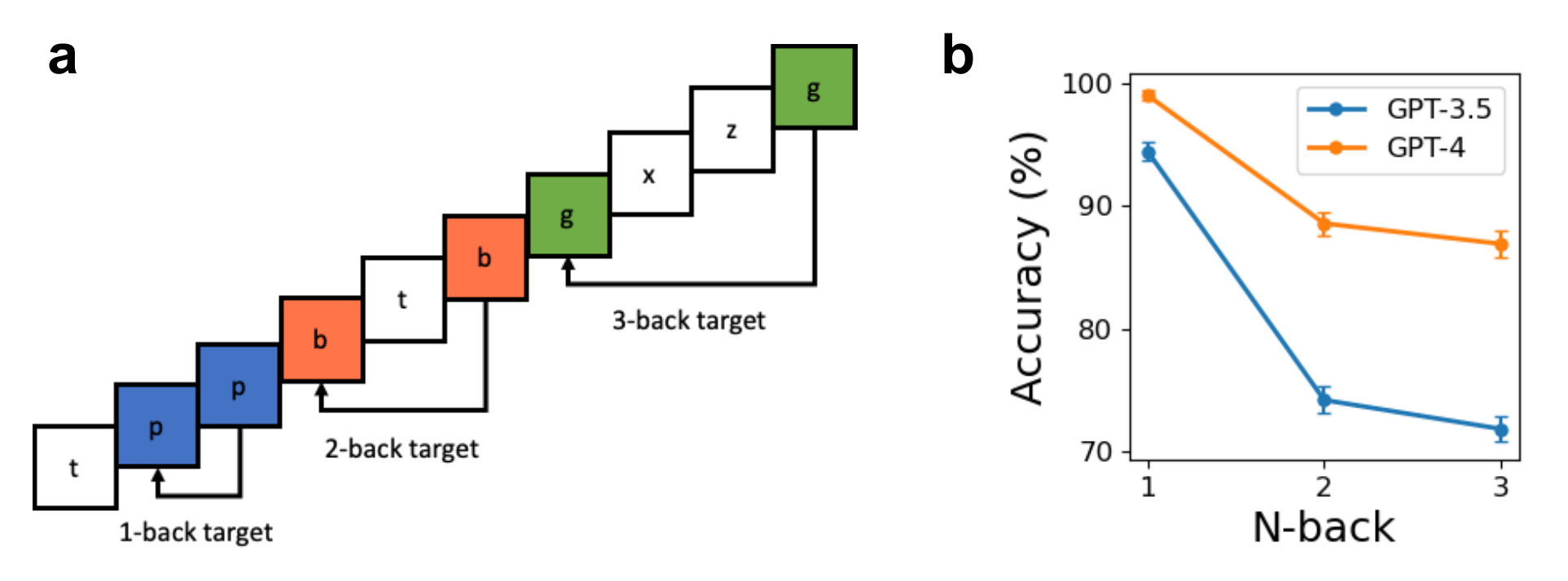}
    \end{center}
    \caption{
           \textbf{(a)}: \textit{N}-back task schematic. Participants (humans or LLMs) are instructed to give a response (humans: press a button; LLMs: output "\texttt{m}") when the current letter is matched with the letter \textit{N} step(s) ago, and withhold responses (humans: do nothing; LLMs: output "\texttt{-}") if it's a nonmatch. \textit{N} is fixed for a given task sequence, and here we put $\{1, 2, 3\}$-back in the same schematic for illustration purposes only. \textbf{(b)}: performance of GPT-3.5 and GPT-4 on this task, reproduced from results in~\cite{gong2024working}. Error bars represent $\pm1$ standard error of the mean.
    }
    \label{fig:n_back}
\end{figure*}

In cognitive science, working memory is defined as the ability of humans to temporarily maintain and manipulate task-relevant information for flexible behaviors~\cite{baddeley1992working}. Recent advancements in Transformer-based LLMs have sparked interest in  evaluating their cognitive abilities, including working memory capacity~\cite{gong2024working}. By designing multiple variants of \textit{N}-back tasks (Figure~\ref{fig:n_back}a)~\cite{kirchner1958age,kaneRolePrefrontalCortex2002} and employing different instructional strategies, it has been found that LLMs consistently perform worse as \textit{N} increases (Figure~\ref{fig:n_back}b), which is reminiscent of the capacity limit of human working memory \cite{cowan2001magical,oberauerWhatLimitsWorking2016,wilhelmWhatWorkingMemory2013}.

However, due to the black-box nature of LLMs, we still lack mechanistic insights as to why the observed capacity limit would emerge, especially given the fact that the length of \textit{N}-back task sequences (e.g., 24 letters in~\cite{gong2024working}) is well within the context length of these models~\cite{pawar2024and}. To answer this question, we were inspired by the executive attention theory~\cite{engle_kane_tuholski_1999,engleWorkingMemoryCapacity2002,Engle2003ExecutiveAW} in human working memory research. The executive attention theory proposes that working memory requires executive attention to maintain access to information in the face of interference, suggesting that it is the scarcity of attentional resources~\cite{lennie_2003,lindsay2020attention}, but not memory storage itself, that is responsible for working memory capacity limits. In Transformer-based LLMs, the self-attention mechanism computes the importance of each element in the input sequence relative to other elements. While this approach allows the model to focus on relevant information, as \textit{N} increases in the \textit{N}-back task, it could be increasingly hard to maintain focus between distant positions. Therefore, we hypothesize that self-attention might be the cause of working memory capacity limits in Transformer-based models.

In the current study, we train causal Transformers on \textit{N}-back tasks and observe that as \textit{N} increases, the model presents a decline in its prediction accuracy. We further find that the prediction accuracy at position $i$ is positively correlated with the attention score at position $i-N$. Furthermore, the model's performance is negatively correlated with the total entropy of the attention score matrix. Our findings suggest that model's inability to aggregate most of its attention to the target position leads to the decline in its prediction accuracy as \textit{N} increases.

\section{Methods}

\paragraph{Dataset.} We use the same procedure described by Gong et al.~\cite{gong2024working} to generate a dataset of \textit{N}-back tasks consisting of task sequences and correct answers. Each task sequence contains 24 letters sampled from an alphabet commonly used in the behavioral literature (``\texttt{bcdfghjklnpqrstvwxyz}"), and the correct answers always consist of 8 matches and 16 nonmatches, mimicking the setup in some human studies. For $N \in \{1, 2, 3, 4, 5, 6\}$, we generate 800 sequences for training and 200 sequences for testing, while our analyses mostly focus on $N \in \{1, 2, 3\}$ to compare with previous studies.

\paragraph{Model.} We use vanilla Transformers in order to facilitate interpretability, as done in prior work aiming to better understand computations in Transformers in more controlled task settings~\cite{elhage2022toy, li2023representations}. We mainly focus our analysis on a causal Transformer containing one decoder layer with only one attention head (Figure~\ref{fig:architecture} in Appendix), although we also test a few architectural variants in the number of decoder layers ($\mathsf{L}$) and number of attention heads per layer ($\mathsf{H}$) for comparisons (see Section \ref{Results} for details). The decoder layer contains masked self-attention so that for each position in the sequence the model can only attend to the current and previous positions. We choose to omit multi-layer feed-forward networks (FFNs) and layer normalization in the original Transformer model to examine the role of self-attention directly without interference from complex internal transformations introduced by FFNs and layer norms. The decoder layer is followed by an unembedding layer to project the decoder outputs to two logits (representing match and nonmatch) for each position.

\paragraph{Training and Evaluation.} We train 50 independent models for each \textit{N}. We choose to train each model for 10 epochs because empirically the model converges after around 10 epochs of training (see Figure~\ref{fig:training loss} in Appendix for details). Cross-entropy loss is computed between the output logits and the correct answers at each position.

\section{Results}
\label{Results}

\begin{figure*}[ht]
    \begin{center}
        \includegraphics[width=0.99\linewidth]{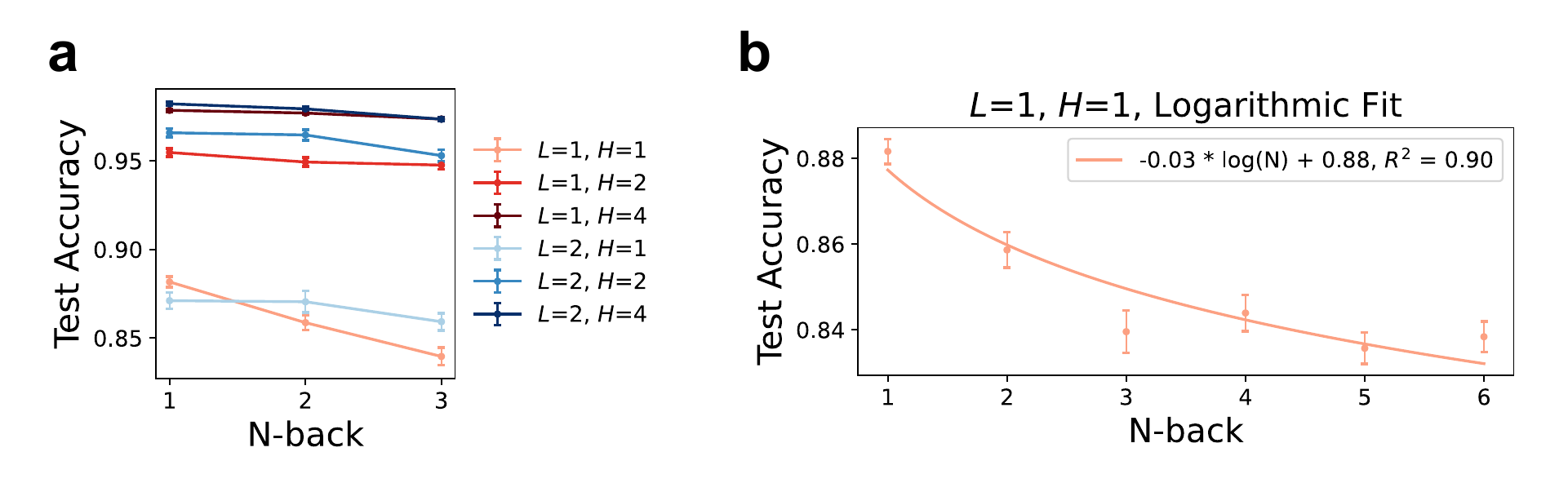}
    \end{center}
    \caption{
           \textbf{(a)}: \textit{N}-back task performance of Transformers with different number of decoder layers and attention heads per layer. \textbf{(b)}: for the 1-layer 1-head Transformer model, task performance drops logarithmically as \textit{N} increases. Error bars represent $\pm1$ standard error of the mean.
    }
    \label{fig:performance}
\end{figure*}

\paragraph{Model accuracy decreases as \textit{N} increases.}

For $\mathsf{L} \in \{1,2\}$ and $\mathsf{H} \in \{1, 2, 4\}$, we train models on the \textit{N}-back task (Figures~\ref{fig:performance}a) and find a significant decline in model performance as \textit{N} increases for the 1-layer 1-head model (Kruskal-Wallis test: H-statistic = 38.517, \textit{p} < .001, $\epsilon^2$ = 0.248; see Table~\ref{table:nback_comparison} in Appendix for post-hoc comparisons using Mann-Whitney U tests\footnote{We use nonparametric Kruskal-Wallis and Mann-Whitney tests instead of \textit{F} and \textit{t} tests because the data do not conform to the assumptions of parametric tests (normality and homogeneity of the variance).}). To further confirm this pattern, we extend the task to \textit{N} = 6, and find a significant logarithmic decline in the test accuracy as \textit{N} increases (Figure~\ref{fig:performance}b). For models with a larger $\mathsf{L}$ or $\mathsf{H}$, most of them achieved over $95\%$ accuracy on all \textit{N}-back tasks. However, they still present slight declines in test accuracy as \textit{N} increases, suggesting that the working memory capacity limit does exist in the nature of transformer models.


\begin{figure*}[ht!]
    \begin{center}
        \includegraphics[width=0.99\linewidth]{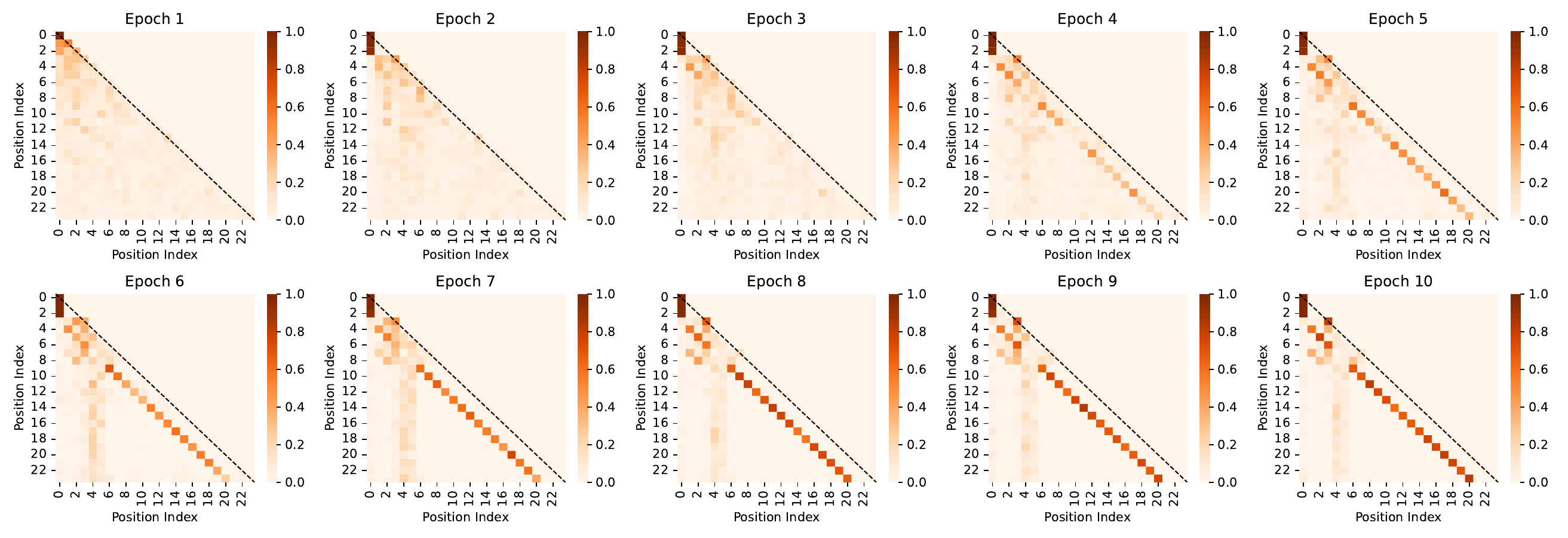}
    \end{center}
    \caption{
           the model learns to attend target locations over training epochs. Here we show attention maps of a 1-layer 1-head Transformer model trained on the 3-back task as an example. See Appendix for attention maps in the 1-back and 2-back tasks.
    }
    \label{fig:weights}
\end{figure*}

\paragraph{Attention scores during training reflect the trajectory of learning.}

To investigate how the self-attention mechanism influences model performance, we visualize attention maps after each training epoch (Figures~\ref{fig:weights},~\ref{fig:1-back attention maps} and~\ref{fig:2-back attention maps}). For each position, we also plot the trajectory of attention scores over training epochs (Figures~\ref{fig:1-back training trajectory},~\ref{fig:2-back training trajectory}, and~\ref{fig:3-back training trajectory}) to see with more granularity how the model learns to perform the task. Starting with almost uniformly distributed attention scores in each row, attention scores gradually aggregate to a line corresponding to the \textit{N}-back positions. For each position in the sequence, attention scores gradually aggregate to the \textit{N}-back position over training epochs and attention scores converge faster for positions earlier in sequence (Figures~\ref{fig:1-back training trajectory},~\ref{fig:2-back training trajectory}, and~\ref{fig:3-back training trajectory}). This shows that the Transformer model learns to master the N-back task by increasing the attention score between the current position and the \textit{N}-back position.

\begin{figure*}[ht]
    \begin{center}
        \includegraphics[width=0.99\linewidth]{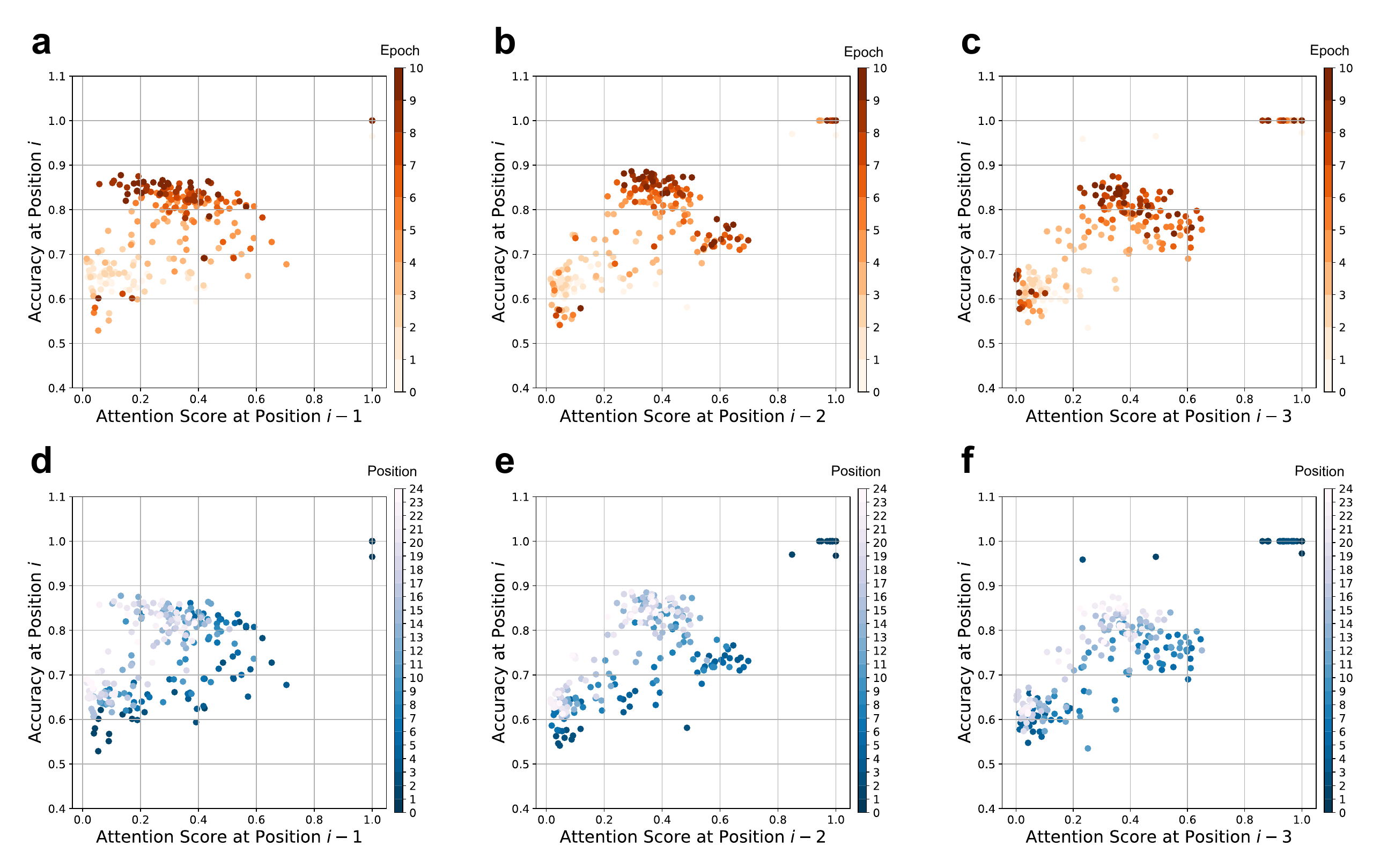}
    \end{center}
    \caption{
           \textbf{(a)}-\textbf{(c)}: the relationship between test accuracy at position $i$ and the attention score at position $i - N$ for the 1-layer 1-head Transformer model. Different colors represent different training epochs each dot belongs to. \textbf{(d)}-\textbf{(f)}: the relationship between test accuracy at position $i$ and the attention score at position $i - N$ for the 1-layer 1-head Transformer model, but here different colors indicate different positions in the task sequence.
    }
    \label{fig:acc_vs_weights}
\end{figure*}

\paragraph{Attention score at position $i-N$ increases with test accuracy at position $i$.}
To further investigate the relationship between attention scores and test accuracy, we plot accuracy at position $i$ against the attention score at the position $i-N$ over training epochs ($i \in \llbracket 1, 24 \rrbracket$, $N \in \{1, 2, 3\}$). The accuracy at position $i$ is defined as the percentage of the model making a correct prediction at position $i$. Over training epochs, we find that the attention score at position $i-N$ increases along with the accuracy at position $i$ (Figure~\ref{fig:acc_vs_weights}a-c), and this is particularly observable for a large $N$ ($N \geq 2$). We reason that in order to produce an accurate prediction at position $i$, the Transformer model needs to learn to put most attention on the $i-N$ position and reduce dispersion of attention to other positions. To better visualize dispersion of attention scores across positions, we use the same data in Figure~\ref{fig:acc_vs_weights}a-c but assign colors to the dots according to which position each dot belongs to (Figure~\ref{fig:acc_vs_weights}d-f). This reveals a clear pattern that attention scores get dispersed at later locations, suggesting that more interference is caused when there are more preceding positions.

\paragraph{Total entropy of attention scores increases as \textit{N} increases.}
Building up from the results above, we take a step further to investigate the overall characteristic of attention scores as \textit{N} increases. 
\newline
To measure the dispersion of attention scores for each \textit{N}, we define the total entropy $H_N$ of each attention score matrix $A \in \mathbb{R}^{24 \times 24}$ as:

\begin{equation}
        H_N(A) = -\sum_{i=1}^{24} \sum_{j=1}^{i} A_{i,j} \log\left(A_{i,j}\right) 
\end{equation}
where
\begin{equation}
     A_{i,j} = \text{Softmax}
     (\frac{QK^T}{\sqrt{d_k}})_{i,j}
\end{equation}

The entropy $H_N$ is well-defined as $\{A_{i,1}, A_{i,2}, ..., A_{i,i}\}$ gives a probability distribution with $\sum_{j=1}^{i} A_{i,j} = 1$ thanks to the Softmax function and causal masking.

\begin{wrapfigure}{r}{0.3\textwidth}
    \centering
    \includegraphics[width=0.3\textwidth]{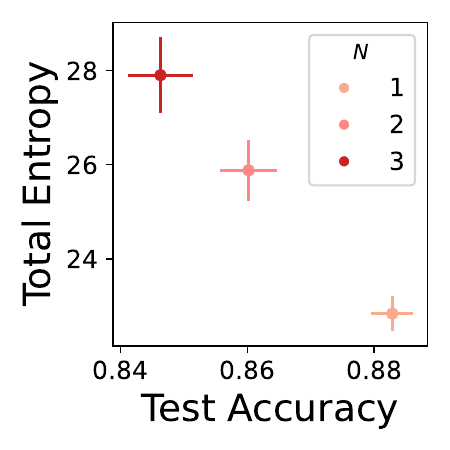}
    \caption{$H_N$ increases as the test accuracy decreases with larger \textit{N}. Error bars represent $\pm1$ standard error of the mean.}
    \label{fig:entropy}
\end{wrapfigure}

For the 1-layer 1-head model, we find that $H_N$ increases as \textit{N} increases, leading to the decrease in test accuracy (Figure~\ref{fig:entropy}). We infer that as \textit{N} increases, it gets harder for the model to learn to attend to the \textit{N}-back letter and the model is less confident about which letter is important, leading to higher entropy and lower accuracy. The fact that large values of $N$ require more structured attention weights (small entropy) to generalize in the \textit{N}-back task is consistent with previous studies on representational learning theory~\cite{liu2022towards}.

\section{Discussion}

The current study provides important insights for the mechanistic interpretability of working memory capacity limits observed in Transformer-based LLMs~\cite{gong2024working}. The self-attention mechanism is critical for the model to achieve good performance in the \textit{N}-back task, but also limits its capacity on the other hand. This is analogous to the mechanism of selective attention in the human brain, which prioritizes relevant information and filter out the rest to ensure effective task performance, but also restricts our information processing by imposing neural and cognitive bottlenecks~\cite{desimone1995neural}. Future work should explore a more formal mathematical proof as to why capacity limits might naturally emerge in complex intelligent systems~\cite{frankland2021no,xie2023natural}.

Although it is still unclear how selective attention in the human brain works at the algorithmic level, we can possibly draw inspirations from how the brain trades off between the amount vs. precision of the information being processed and design better model architectures with enhanced working memory capacity, which could in-turn lead to more powerful model capabilities in reasoning and problem solving~\cite{jaeggiImprovingFluidIntelligence2008,halfordSeparatingCognitiveCapacity2007}.

Note that the current study focuses on a very simplified version of the Transformer model, so it is not straightforward to draw direct comparisons with pre-trained LLMs such as those evaluated by Gong et al.~\cite{gong2024working}. It is thus important for future research to investigate how the complexity of the model architecture and the number of learnable parameters in the model would influence task performance. In addition, varying the amount of training data and the specific hyperparameters used during training would also be crucial for understanding model behaviors in finer detail.

\newpage
\bibliographystyle{plain}
\sloppy
\bibliography{references}

\newpage
\appendix
\section*{Appendix}
\begin{figure*}[h!]
    \begin{center}
        \includegraphics[width=0.5\linewidth]{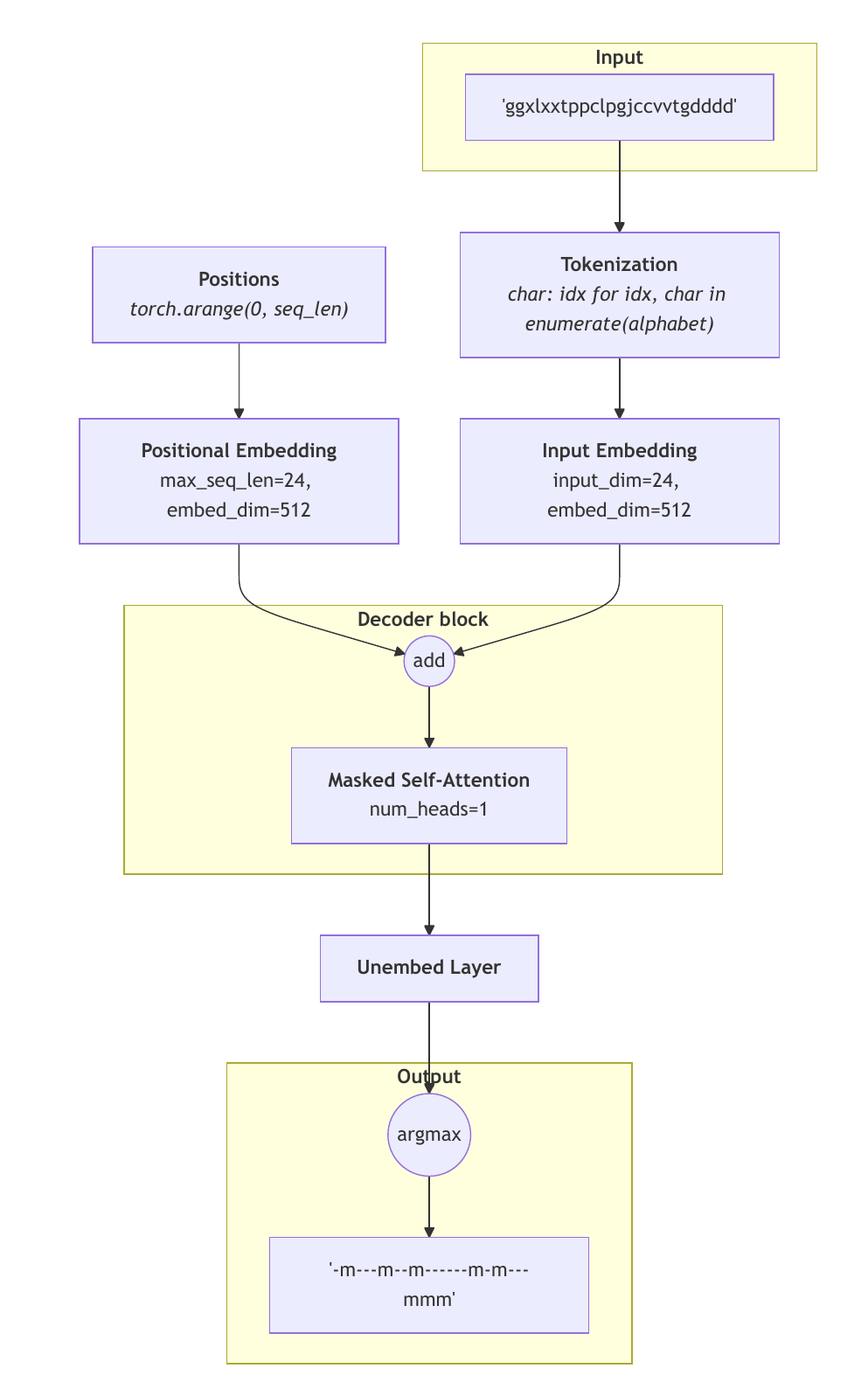}
    \end{center}
    \caption{
           The architecture of the 1-layer 1-head Transformer.
    }
    \label{fig:architecture}
\end{figure*}

\begin{figure*}[h!]
    \begin{center}
        \includegraphics[width=0.8\linewidth]{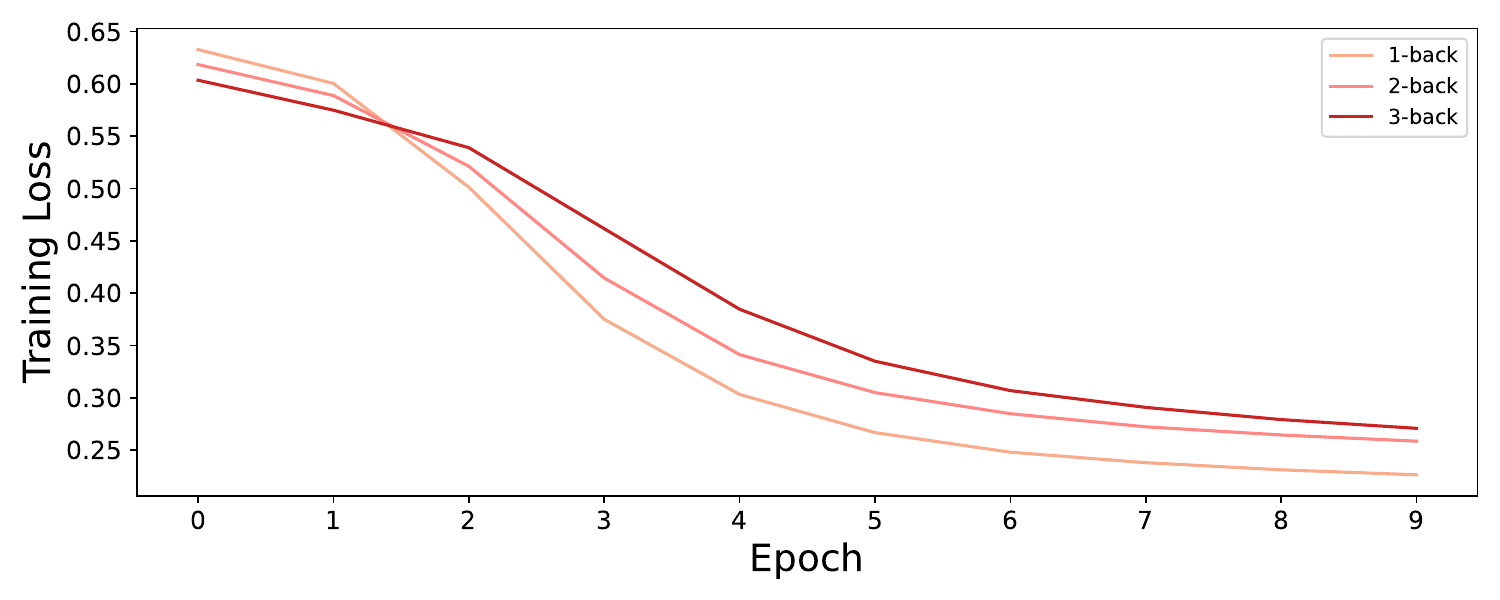}
    \end{center}
    \caption{
           Training loss of the 1-layer 1-head Transformer converges after 10 epochs.
    }
    \label{fig:training loss}
\end{figure*}

\begin{table}[ht!]
    \vspace{-3mm}
    \centering
    \caption{Post-hoc Mann-Whitney \textit{U} test results for the 1-layer 1-head model.}
    \begin{tabular}{lrrr}
    \toprule
        \textit{N}-back & \textit{U} & \textit{p} & r \\ 
    \midrule
        1 vs 2 & 1825.0000 & 0.0002 & -0.4600 \\
        1 vs 3 & 2096.0000 & 0.0000 & -0.6768 \\
        2 vs 3 & 1665.0000 & 0.0128 & -0.3320 \\ 
    \bottomrule
    \end{tabular}
    \label{table:nback_comparison}
\end{table}

\begin{figure*}[h!]
    \begin{center}
        \vspace{-3mm}
        \includegraphics[width=0.99\linewidth]{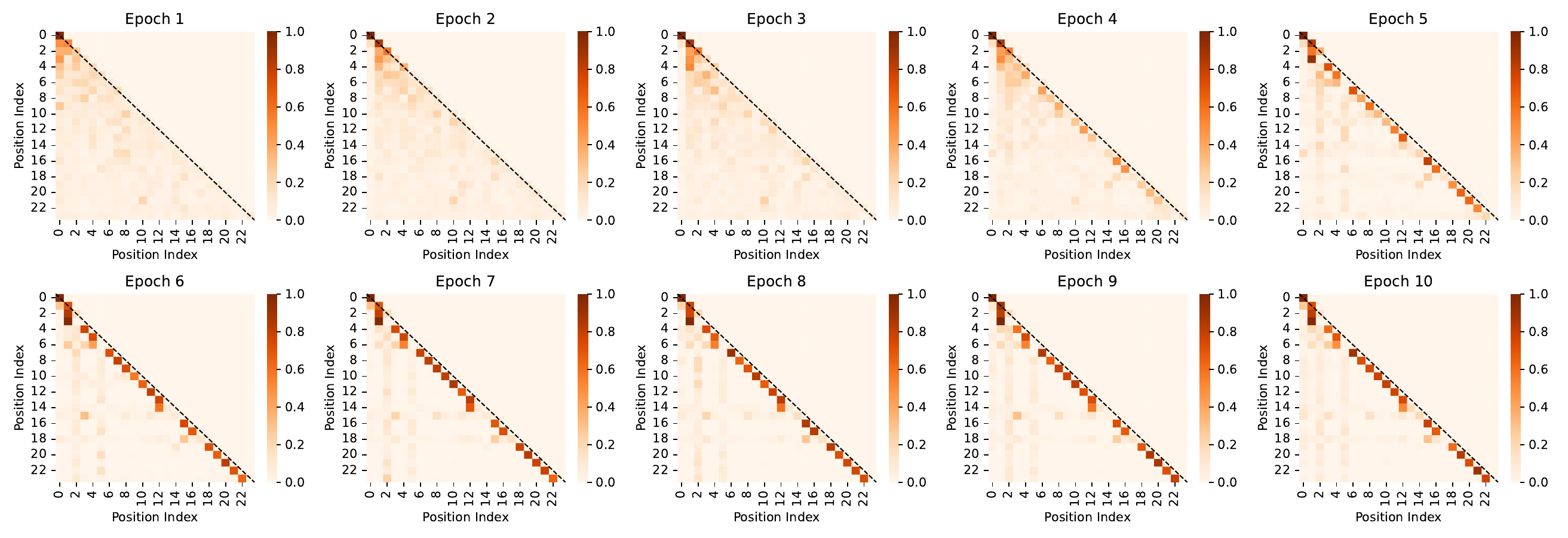}
    \end{center}
    \vspace{-3mm}
    \caption{
    Attention maps over training epochs for a 1-layer 1-head Transformer trained on the 1-back task.
    }
    \vspace{-3mm}
    \label{fig:1-back attention maps}
\end{figure*}

\begin{figure*}[h!]
    \begin{center}
        \includegraphics[width=0.99\linewidth]{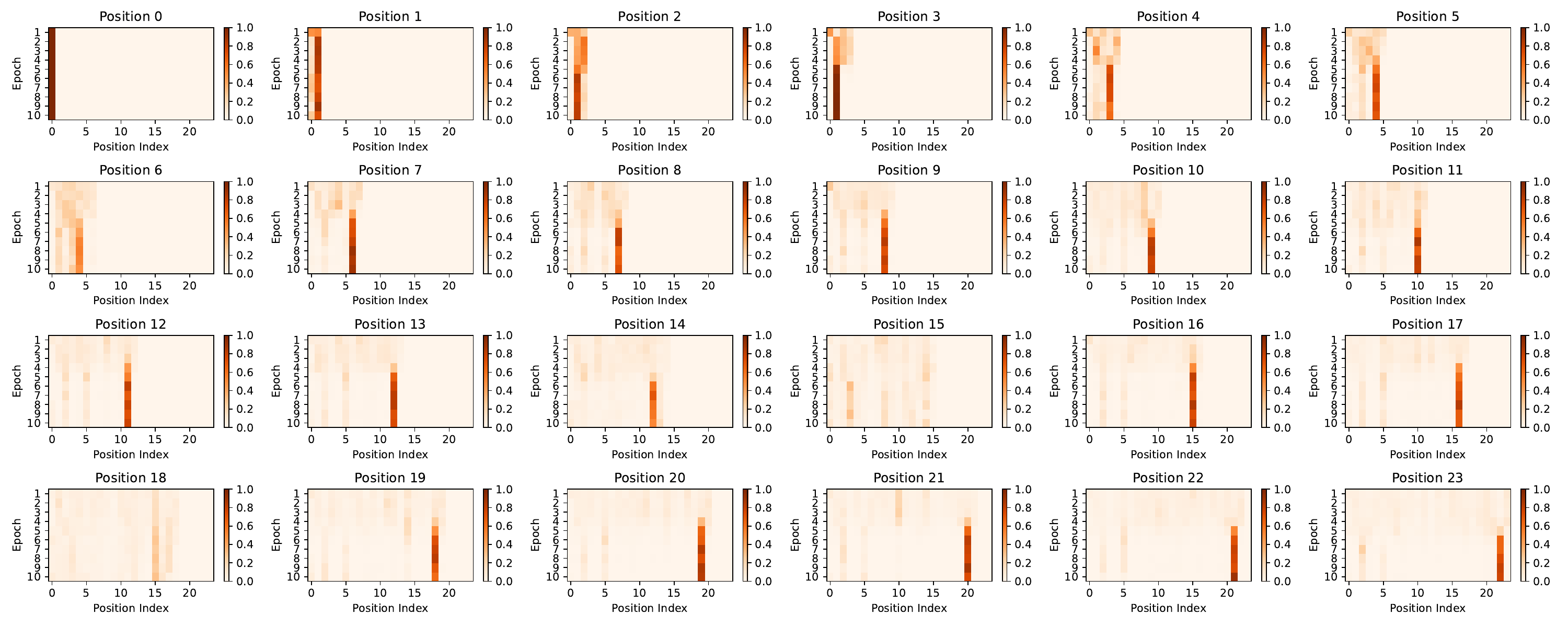}
    \end{center}
    \caption{Training trajectory of attention scores over 10 epochs for the 1-back task.          
    }
    \label{fig:1-back training trajectory}
\end{figure*}

\begin{figure*}[h!]
    \begin{center}
        \includegraphics[width=0.99\linewidth]{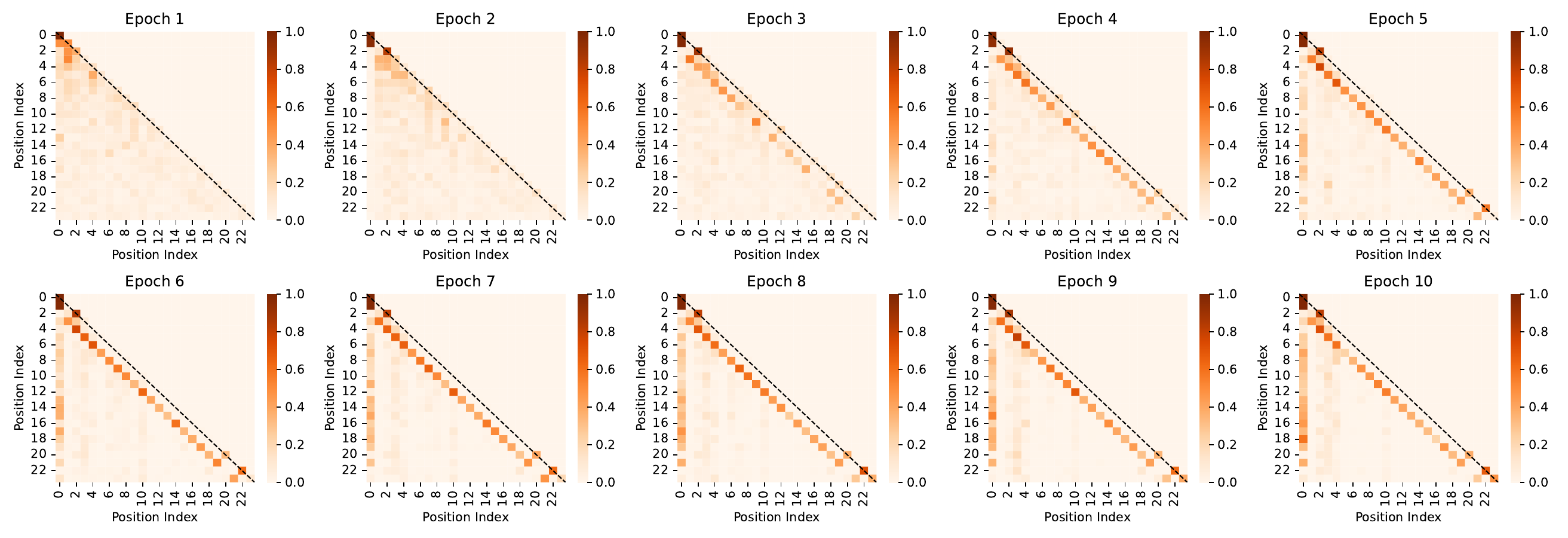}
    \end{center}
    \caption{    
    Attention maps over training epochs for a 1-layer 1-head Transformer trained on the 2-back task.
    }
    \label{fig:2-back attention maps}
\end{figure*}

\begin{figure*}[ht]
    \begin{center}
        \includegraphics[width=0.99\linewidth]{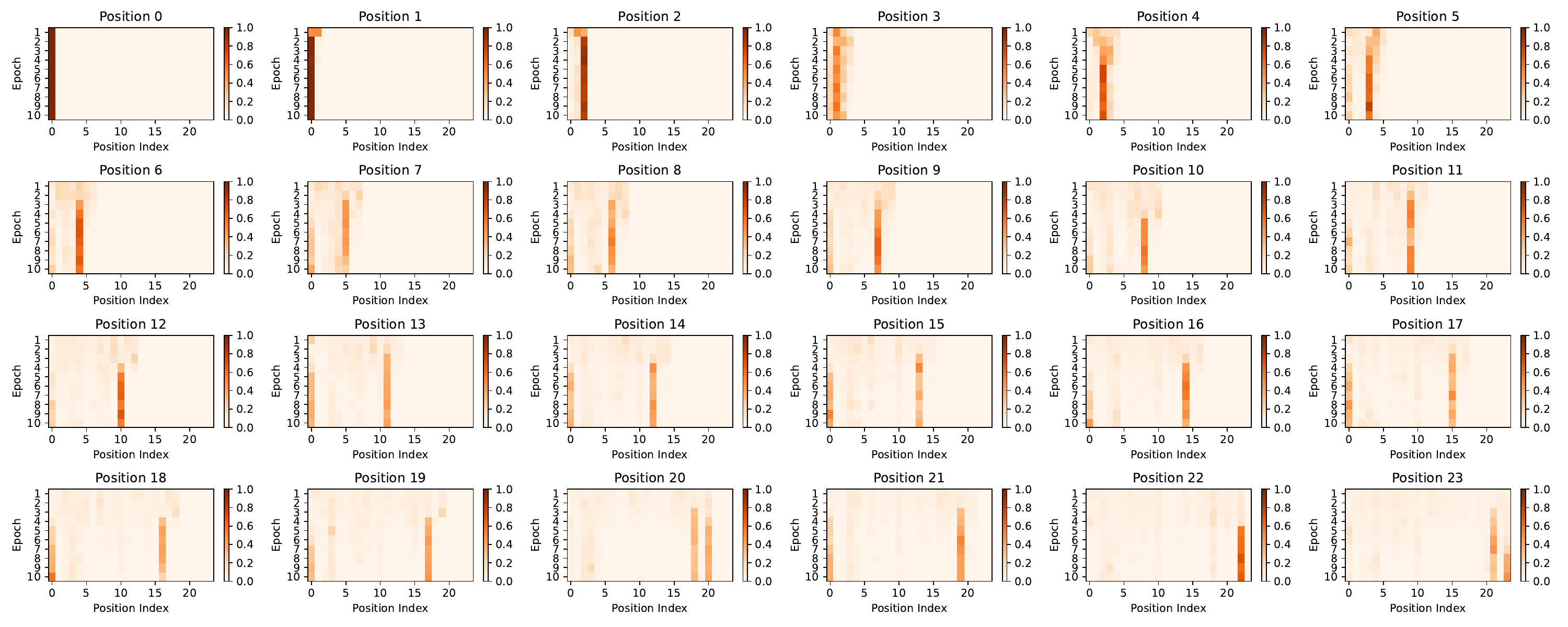}
    \end{center}
    \caption{Training trajectory of attention scores over 10 epochs for the 2-back task.   
    }
    \label{fig:2-back training trajectory}
\end{figure*}

\begin{figure*}[ht]
    \begin{center}
        \includegraphics[width=0.99\linewidth]{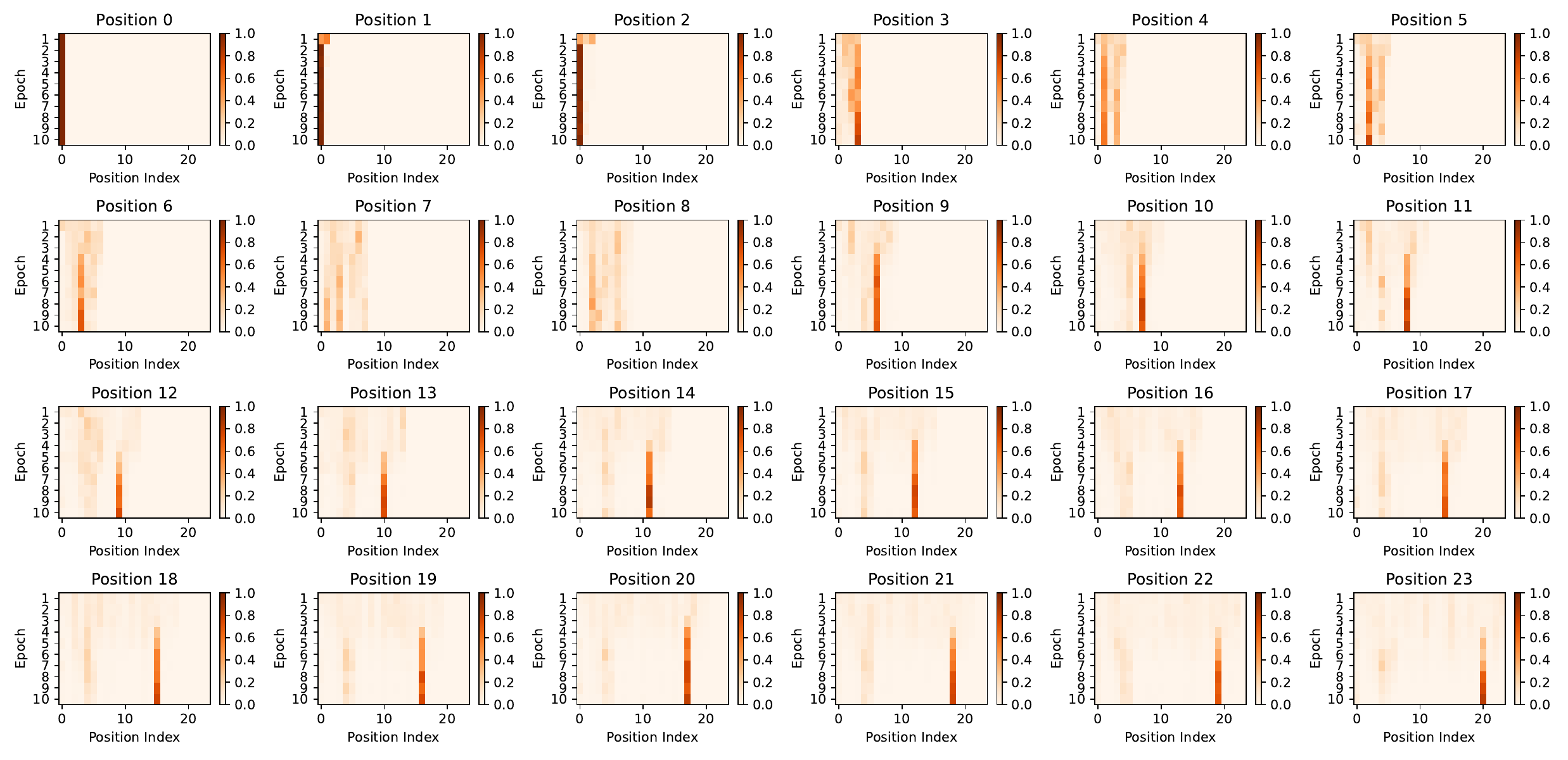}
    \end{center}
    \caption{Training trajectory of attention scores over 10 epochs for the 3-back task.        
    }
    \label{fig:3-back training trajectory}
\end{figure*}

\end{document}